\documentclass{article}
\usepackage{spconf,amsmath,graphicx}
\usepackage{multirow}
\usepackage{multicol}
\usepackage{booktabs}
\usepackage{todonotes}
\usepackage{mathtools}
\usepackage{xcolor}
\usepackage[noadjust]{cite}
\usepackage[latin1,utf8]{inputenc}
\usepackage[T1]{fontenc}
\usepackage{hyperref}
\usepackage{caption}
\usepackage{subcaption}

{}
\newcommand{\etal}{\textit{et al}.}

\title{PerceptNet: A human visual system inspired neural network for estimating perceptual distance}

\name{Alexander Hepburn$^{\star}$\thanks{This work was partially funded by EPSRC grant EP/N509619/1 and MINECO grant DPI2017-89867.} \quad Valero Laparra$^{\dagger}$ \quad Jes\'{u}s Malo$^{\dagger}$ \quad Ryan McConville$^{\star}$ \quad Raul Santos-Rodriguez$^{\star}$}
\address{$^{\star}$ Department of Engineering Mathematics, University of Bristol \\
$^{\dagger}$Image Processing Lab, Universitat de Valencia}
\begin{document}
%
\maketitle
\begin{abstract}
Traditionally, the vision community has devised algorithms to estimate the distance between an original image and images that have been subject to perturbations. Inspiration was usually taken from the human visual perceptual system and how the system processes different perturbations in order to replicate to what extent it determines our ability to judge image quality. While recent works have presented deep neural networks trained to predict human perceptual quality, very few borrow any intuitions from the human visual system. To address this, we present PerceptNet, a convolutional neural network where the architecture has been chosen to reflect the structure and various stages in the human visual system. We evaluate PerceptNet on various traditional perception datasets and note strong performance on a number of them as compared with traditional image quality metrics. We also show that including a nonlinearity inspired by the human visual system in classical deep neural networks architectures can increase their ability to judge perceptual similarity.
Compared to similar deep learning methods, the performance is similar, although our network has a number of parameters that is several orders of magnitude less.
\end{abstract}
\section{Introduction}
\label{sec:intro}
The human visual system's ability to compare images through a variety perturbations is still unparalleled. Within machine learning and computer vision, \emph{perception} has become increasingly relevant, and many metrics have attempted to capture characteristics of the human perceptual system in order to replicate the ability to perceive differences between images. Despite being a well established field, an often overlooked aspect is where the foundations of image processing originate; the human visual perceptual system. In the past, subjective image quality metrics were proposed following two principles: the \emph{visibility of errors} derived from psychophysical models~\cite{Watson93,Teo94,Malo02,Laparra10}, or the preservation of perceptual \emph{structural similarity} as in SSIM and variants~\cite{wang2003multiscale,wang2004image,zhang2011fsim}. 
More recent efforts has been focused on training neural networks to distinguish between patches of images~\cite{zhang2018unreasonable}. Although these networks are shown to be successful in datasets containing a wide variety of perturbations, the network structure takes no inspiration from the human visual system. These networks also contain millions of parameters and are often difficult to interpret. In fact, recently it has been shown that blind fitting of architectures which are not properly constrained may lead to failures in reproducing fundamental perceptual phenomena~\cite{texture-bias,Martinez19}. 

We propose combining the recent methodologies from deep learning approaches and traditional image quality metrics by constructing a distance where the architecture
takes inspiration from what we understand about the human visual system. In this paper we present PerceptNet, a carefully constructed network that has been trained on a limited set of perturbations and has an ability to generalise to perturbations in other datasets. PerceptNet outperforms classical measures and in traditional image quality databases and performs similarly to deep learning measures despite having two orders of magnitude less parameters.

\section{Related Work}
\label{sec:related}
Image quality metrics (IQMs) have long been relying on our understanding of the human visual perceptual system. Methods based on assessing the \emph{visibility of errors} apply models of the psychophysical response to the original image and to the distorted image, and then compute Euclidean distances in the transformed domain. These response models have always been cascades of linear+nonlinear layers, mainly wavelet-like filters followed by divisive normalisation saturations.
The difference between old implementations of this idea~\cite{Watson93,Teo94,Malo02,Laparra10} and newer ones~\cite{laparra2016,martinez2018derivatives} is the biological sophistication of the models and the way they are optimised.
Models based on \emph{structural similarity} such as SSIM~\cite{Wang2004} and its variants MS-SSIM and FSIM~\cite{wang2003multiscale,zhang2011fsim} check the integrity of the statistics of the distorted image. However, it has been shown error visibility models may be as adaptive as structural similarity~\cite{Laparra10}. Therefore, current versions of error visibility models based on normalised laplacian pyramids (NLAPD)~\cite{laparra2016,laparra2017perceptually} clearly outperform structural similarity methods whilst also shown to be effective at enforcing perceptual quality in image generation~\cite{hepburn2019enforcing}.
%
The linear+divisive normalisation layer can also be formulated as a convolutional+nonlinear layer in CNNs, called Generalised Divisive Normalisation  (GDN)~\cite{balle2015density}. 

CNNs have previously been used for full reference image quality assessment~\cite{bosse2017deep}, although the proposed architectures do not reflect the structure of the human visual system. A popular full reference metric is the Learned Perceptual Image Patch Similarity (LPIPS)~\cite{zhang2018unreasonable}. LPIPS utilises architectures trained to classify images on ImageNet~\cite{imagenet} as feature extractors. A weighted importance vector is learned and a combination of spatial average and $\ell_2$ distance is used to compute the perceptual distance. An alternative is to train using random initialisations and the Berkeley-Adobe Perceptual Patch Similarity (BAPPS) dataset. BAPPS contains traditional perceptual distortions, convolution-based perturbations and a combination of both. In traditional human judgement experiments, the observer is presented with an original image and two distorted images and is asked to select the distorted image that is most similar with the original. BAPPS contains only the fractional preference for each combination of two images and, as such, the distance output from the networks must be transformed into a preference score. A network $\mathcal{G}$ containing two fully connected layers is used, which takes as input the distances from the original to both distorted images and outputs a predicted preference.
Interestingly, it is known that convolution neural networks trained on ImageNet have a texture bias which contradicts what we know about the human visual perceptual system~\cite{texture-bias}. 
Similar departures from the desired perceptual behaviour have been also reported when the training set is not appropriate and the architecture is not properly constrained~\cite{Martinez19}.

\section{A Perceptually Constrained Architecture}
\label{sec:network}
To address the aforementioned shortcomings of current deep learning approaches, we devise an architecture, PerceptNet, for our proposed networks following the program suggested in~\cite{Carandini12}: a cascade of canonical \emph{linear filters + divisive normalisation} layers that perform a series of perceptual operations in turn simulating the retina - LGN - V1 cortex pathway~\cite{martinez2018derivatives}.

\begin{figure}[t]
    \centering
    \includegraphics[width=\columnwidth]{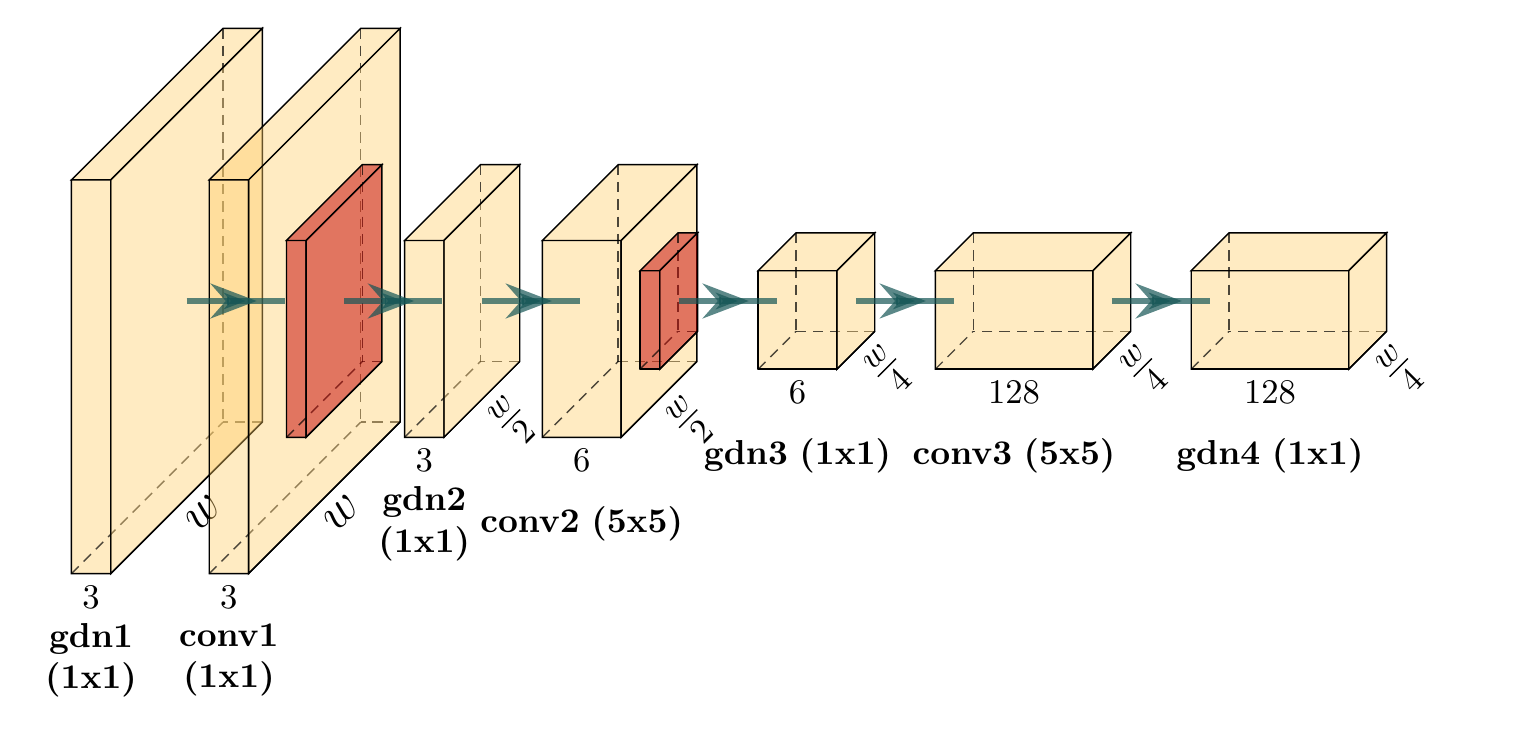}
    \caption{A diagram displaying the structure of PerceptNet. Each stage has a corresponding process in the human visual system: gamma correction $\xrightarrow{}$ opponent colour space $\xrightarrow{}$ Von Kries transform $\xrightarrow{}$ center-surround filters $\xrightarrow{}$ LGN normalisation $\xrightarrow{}$ Orientation sensitive and multiscale in V1$\xrightarrow{}$ divisive normalisation in V1. $(a\times a)$ denotes the filter size used. gdn1 is applied independent to each channel.}
    \label{fig:my_arch}
\end{figure}

The architecture is depicted in Fig.~\ref{fig:my_arch}. Firstly, we use GDN to learn Weber-like saturation~\cite{Fairchild05} at the RGB retina. Then, we learn a linear transformation to an opponent colour space, analogue to the achromatic, red-green, yellow-blue colour representation in humans~\cite{Fairchild05}.
This linear transform is subsequently  normalised again using GDN to learn a chromatic adaption process similar to Von-Kries~\cite{Abrams07}.
Afterwards, spatial convolutions are allowed to learn center-surround filters as in LGN~\cite{Shapley11}, which are known to have nonlinear GDN-like behaviour~\cite{Malo02,laparra2016,martinez2018derivatives}. Finally, we include a new convolution+GDN stage to account for the wavelet-like filters at V1 cortex and the divisive normalisation~\cite{Watson97,Laparra10}. This domain replicates the representation at the end of the primary visual cortex, where most of the information is contained in various orientation sensitive edge detectors whilst preserving a map of spatial information.

The network is trained to maximise the Pearson correlation, $\rho$, between the mean opinion score (MOS) and the $\ell_2$ distance of the two images in the transformed domain:
\begin{equation} \label{corr-loss}
    \max_f \rho(||f(x) - f(d(x))||_2, y),
\end{equation}
where $f(x)$ is the transformation of the network (from RBG space to the more informative perceptual space), $x$ is the reference image, $d(x)$ is the distorted image and $y$ is the corresponding MOS calculated from human observer experiments.

A number of properties are recognisable in the way that humans process images; one being a focus on medium frequency in the receptive field~\cite{Malo02}. The contrast sensitivity function of the spatial standard observer (SSO) models this behaviour. The SSO model is used to judge perceptual distance between two contrast patterns and tends to also focus on medium frequencies as a result. Our network captures these characteristics as they are intrinsically linked to judging human perceptual distance.

\section{Experiments}
It is our aim to find a representation that informs us of the overall perceptual quality, generalising to distortions not seen during the training phase. To this end, we use the TID2008~\cite{tid2008-data} dataset for training. It contains 17 distortions, with $1428$ distorted images and corresponding MOS. Our code and models are publicly available \footnote{\url{https://github.com/alexhepburn/perceptnet}}.
We evaluate the network on multiple perceptual datasets; TID2013~\cite{tid2013-data}, CSIQ~\cite{csiq-data}, LIVE~\cite{live-data} and BAPPS~\cite{zhang2018unreasonable}. A simple description of the datasets can be seen in Table~\ref{tab:datasets}. Both TID2013 and BAPPS contain distortions that are not present in the TID2008 dataset.
The BAPPS dataset is slightly different in that it does not report MOS. The dataset also contains interesting distortions such as convolutional neural network (CNN) based distortions, while the test set contains distortions that are made using other deep learning algorithms, such as super-resolution, video deblurring and colourisation.

We will be comparing PerceptNet with several baselines, namely, the $\ell_2$ distance between reference and distorted image, traditional IQMs like SSIM~\cite{wang2004image}, FSIM~\cite{zhang2011fsim} and MSSIM~\cite{wang2003multiscale}. We also compare against the NLAPD proposed in~\cite{laparra2016}, but we replace the divisive normalisation step and each stage in the pyramid with a generalised divisive normalisation process, where the parameters are optimised using the TID2008 dataset. The main method using deep learning architectures is the LPIPS measure proposed in~\cite{zhang2018unreasonable}. Zhang \etal found that LPIPS AlexNet initialised from scratch trained on the BAPPS dataset performed best on the BAPPS test subset. For LPIPS, \texttt{scratch} denotes that the network was trained from random initialisation, and \texttt{tune} indicates that the network was pretrained on a dataset and fine-tuned to the BAPPS dataset. We also train AlexNet on perceptual datasets as a feature extractor. Importantly, it should be noted that LPIPS AlexNet requires $24.7m$ parameters whereas PerceptNet has $36.3k$ parameters. ImageNet contains millions of images compared to traditional perceptual datasets, which usually have thousands or hundreds of examples. When AlexNet is trained on the traditional perceptual datasets such as TID2008, we use the feature extractor section of the network and disregard the classification layers. We train the network using the correlation loss in Eq.~\ref{corr-loss}. When comparing with LPIPS, it is important to provide comparisons using the test subset of the dataset it was trained on -- the BAPPS dataset.
Although the LPIPS measures are trained using the $G$ network that transforms two distances to a fractional preference, when evaluating the measure, only the main network is used. One measure for evaluating IQMs is two-alternative forced choice (2AFC). This is the percentage of images where the image closest in distance to the reference using the specific measure agrees with the majority of human voters.

\begin{figure}[!h]
     \centering
     \begin{subfigure}[b]{0.49\columnwidth}
         \centering
         \includegraphics[width=\textwidth]{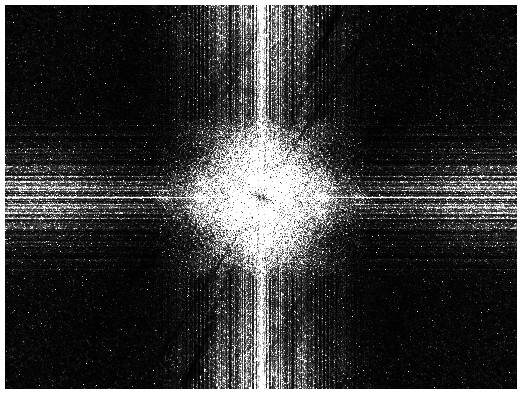}
         \caption{Channel 88}
         \label{fig:receptivefield1}
     \end{subfigure}
     \begin{subfigure}[b]{0.49\columnwidth}
         \centering
         \includegraphics[width=\textwidth]{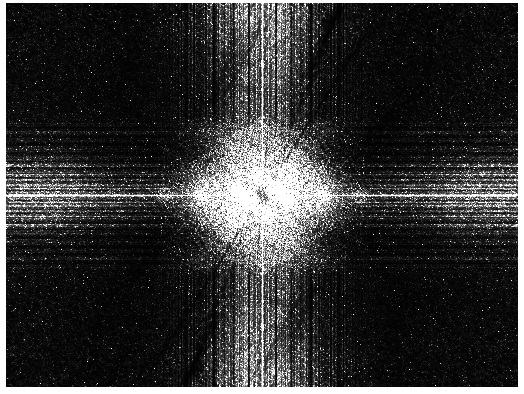}
         \caption{Channel 64}
         \label{fig:receptivefield2}
     \end{subfigure}
     \caption{Receptive field in the Fourier domain for channels that maximise $\ell_2$ distance between reference and an achromatic JPEG2k transmission error distorted image (from Fig.~\ref{fig:output}).}
     \label{fig:receptivefield}
\end{figure}

\begin{table}[!ht]
\centering
    \centering
    \begin{tabular}{@{}lll@{}}
    \toprule
        Dataset & \begin{tabular}[c]{@{}l@{}}Number of\\ Samples\end{tabular} & \begin{tabular}[c]{@{}l@{}}Number of\\ Distortions\end{tabular} \\ \midrule
        TID2008~\cite{tid2008-data} Train& 1428 & 17 \\
        TID2008~\cite{tid2008-data} Test & 272 &  17 \\
        TID2013~\cite{tid2013-data} & 3000 & 24 \\
        CSIQ~\cite{csiq-data} & 899 & 6 \\ 
        LIVE~\cite{live-data}& 982 & 5 \\
        BAPPS~\cite{zhang2018unreasonable} Train & 151.4k & 425 \\ 
        BAPPS~\cite{zhang2018unreasonable} Test & 36.3k & 425+ \\ \bottomrule
    \end{tabular}
    \caption{An overview of the datasets used in the paper.}
    \label{tab:datasets}
\end{table}

\begin{table*}[!h]
\centering
\begin{tabular}{@{}llllll@{}}
\toprule
\multirow{2}{*}{Method} & \multirow{2}{*}{\begin{tabular}[c]{@{}l@{}}Trained\\ On\end{tabular}} & \multicolumn{4}{l}{Pearson Correlation (Spearman Correlation) with MOS} \\ \cmidrule(l){3-6} 
 &  & TID2008 Test & TID2013 & CSIQ & LIVE \\ \midrule
SSIM &  & 0.51 (0.53) & 0.62 (0.60) & 0.77 (0.84) & 0.84 (0.95) \\
MS-SSIM &  & 0.78 (0.80) & 0.78 (0.80) & 0.81 (0.91) & 0.77 (0.97) \\
FSIM$_c$ &  & 0.79 (0.84) & 0.79 (0.81) & 0.82 (0.93) & 0.77 (0.92)\\
NLAPD (with GDN)& TID2008 & 0.81 (0.82) & 0.82 (0.81) & 0.90 (0.92) & 0.88 (0.96) \\
AlexNet (with ReLU) & TID2008 & 0.89 (0.89) & \textbf{0.93} \textbf{(0.91)} & \textbf{0.95} (0.95) & 0.88 (0.94) \\
AlexNet (with GDN) & TID2008 & 0.91 (0.91) & 0.92 \textbf{(0.91)} & 0.94 (0.95) & 0.93 (0.95) \\
PerceptNet & TID2008 & \textbf{0.93 (0.93)} & 0.90 (0.87) & 0.94 \textbf{(0.96)} & \textbf{0.95 (0.98)} \\
LPIPS AlexNet (tune) & \begin{tabular}[c]{@{}l@{}}ImageNet + BAPPS\end{tabular} & 0.74 (0.75) & 0.76 (0.76) & 0.88 (0.93) & 0.85 (0.96) \\
LPIPS AlexNet (scratch) & BAPPS & 0.47 (0.47) & 0.58 (0.57) & 0.72 (0.80) & 0.77 (0.89) \\
PerceptNet (tune) & \begin{tabular}[c]{@{}l@{}}TID2008 + BAPPS\end{tabular} & 0.67 (0.72) & 0.75 (0.76) & 0.81 (0.88) & 0.85 (0.94) \\
PerceptNet (scratch) & BAPPS & 0.56 (0.67) & 0.67 (0.72) & 0.77 (0.84) & 0.80 (0.93) \\ \bottomrule
\end{tabular}
\caption{Traditional IQMs and state-of-the-art approaches evaluated on a variety of datasets. We report the Pearson and Spearman correlations between distances obtained using these methods and the MOS. For methods that are feature extractors (AlexNet, PerceptNet) we took the $\ell_2$ distance between features obtained using the reference and distorted images.}
\label{tab:correlation-results}
\end{table*}

\begin{table*}[!htb]
\centering
\begin{tabular}{lllllllll}
\hline
\multirow{2}{*}{Method} & \multirow{2}{*}{Trained On} & \multicolumn{7}{c}{2AFC Accuracy (\%)} \\ \cline{3-9} 
 &  & Average & \begin{tabular}[c]{@{}l@{}}Trad-\\ itional\end{tabular} & \begin{tabular}[c]{@{}l@{}}CNN\\ Based\end{tabular} & \begin{tabular}[c]{@{}l@{}}Super\\ Res\end{tabular} & \begin{tabular}[c]{@{}l@{}}Video\\ Deblur\end{tabular} & \begin{tabular}[c]{@{}l@{}}Colour-\\ isation\end{tabular} & \begin{tabular}[c]{@{}l@{}}Frame\\ Interp\end{tabular} \\ \hline
LPIPS AlexNet (tune) & ImageNet + BAPPS & 69.7 & 77.7 & 83.5 & 69.1 & 60.5 & 64.8 & 62.9 \\
LPIPS AlexNet (scratch) & BAPPS & 70.2 & 77.6 & 82.8 & 71.1 & 61.0 & 65.6 & 63.3 \\
LPIPS PerceptNet (tune) & TID2008 + BAPPS & 67.8 & 69.4 & 81.3 & 70.6 & 60.9 & 61.9 & 62.6 \\
LPIPS PerceptNet (scratch) & BAPPS & 69.2 & 75.3 & 82.5 & 71.3 & 61.4 & 63.6 & 63.2 \\
AlexNet & TID2008 & 63.2 & 56.1 & 77.4 & 66.1 & 58.6 & 61.6 & 56.2 \\
PerceptNet & TID2008 & 64.9 & 58.1 & 80.5 & 68.3 & 59.6 & 61.6 & 58.2 \\ \hline
\end{tabular}
\caption{Two-alternative forced choice (2AFC) accuracy scores for various architectures, all evaluated on the BAPPS~\cite{zhang2018unreasonable} dataset. The accuracy is the percentage of samples that the method agreed with the majority of human observers.}
\label{tab:afc-scores}
\end{table*}

\begin{figure*}[!htb]
    \centering
    \includegraphics[width=\textwidth]{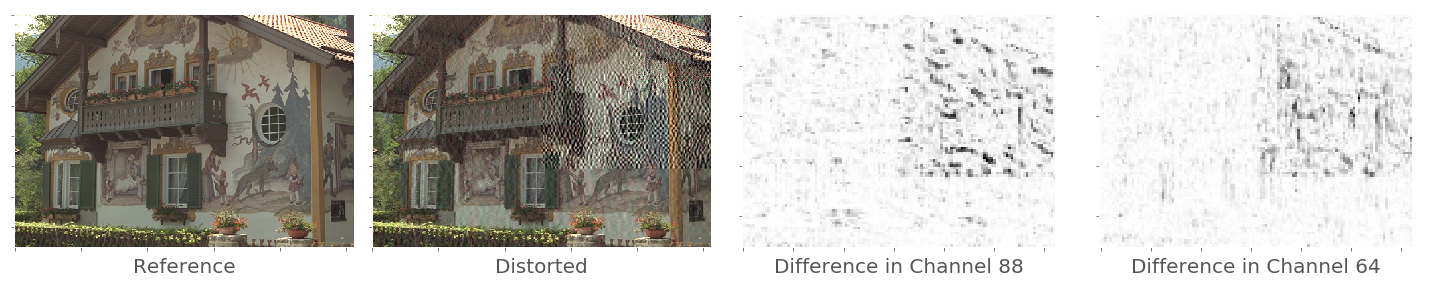}
    \caption{Difference in the output of the network for a reference image and distortion image. The channels shown are those that are the maximum in $\ell_2$ distance between the outputs. Each difference in channels was scaled to $[0, 255]$. The image is from the TID2008 test set and the distortion is the maximum magnitude for JPEG2000 transmission errors.}
    \label{fig:output}
\end{figure*}
Table~\ref{tab:correlation-results} contains Pearson and Spearman correlations between MOS and distances measured using multiple algorithms. The higher the correlations, the more aligned the distance is to how humans judge the distorted images. PerceptNet outperforms all algorithms on the TID2008 Test and LIVE datasets. AlexNet is the best performing algorithm on a dataset with unseen distortions, such as TID2013 but the number of parameters in AlexNet is two orders of magnitude greater than PerceptNet. The networks trained using BAPPS (LPIPS and PerceptNet tune and scratch) perform similar to traditional IQMs.
Replacing the ReLU activation with GDN layers in AlexNet improves correlations on the TID2008 Test and LIVE datasets.

Table~\ref{tab:afc-scores} contains 2AFC scores for multiple networks and datasets. Networks trained on perceptual datasets perform poorly on the BAPPS dataset and networks trained on the BAPPS dataset perform poorly on the perceptual datasets containing less distortions (table \ref{tab:correlation-results}). 
Training AlexNet and PerceptNet on TID2008, and evaluating on BAPPS, leads to similar results despite AlexNet having a larger number of parameters. Using LPIPS with PerceptNet, instead of AlexNet, leads to slightly worse performance when the pretrained networks tuned on BAPPS. Training from scratch on BAPPS, PerceptNet outperforms AlexNet it is able to better generalise to other perceptual datasets.
 
Fig.~\ref{fig:output} shows an example of the output from PerceptNet for a reference and distorted image. Each channel has been scaled to $[0, 255]$. The main difference between the channels are where the distortions have taken place. 
Fig.~\ref{fig:receptivefield} shows the receptive field in the Fourier domain for the corresponding channels (88 and 64). These resemble the Contrast Sensitivity Function of the Spatial Standard Observe in that the channels focus on the mid-frequencies, where humans have maximum sensitivity~\cite{Malo02}.


\section{Conclusion}
We describe a transformation inspired by the different stages in the human visual system that can accurately predict human perceived distance in images subjected to a number of distortions. This transformation is implemented as a deep neural network. We show that this network can generalise to datasets with more distortions than are present in the training set and clearly performs better than traditional IQMs. Importantly, although it has two orders of magnitude less parameters, the performance is similar to AlexNet. Visualising the output of the transformation shows that the perceptual space contains a number of properties that are thought to be present in the human visual system. We also show that substituting ReLU layers with GDN layers in AlexNet increases its ability to judge perceptual similarity.


\clearpage
\bibliographystyle{IEEEbib}
\bibliography{ref}

\end{document}